\documentclass[conference]{IEEEtran}
\usepackage{times}

\IEEEoverridecommandlockouts                              
\usepackage{amsmath}
\usepackage{amssymb}
\usepackage{mathtools}
\usepackage{graphicx}
\usepackage{float}
\usepackage{array}
\usepackage{tikz}
\usepackage{listings}
\usepackage[bookmarks=true]{hyperref}
\usepackage{graphicx}
\usepackage{cite}
\usepackage{dsfont}
\usepackage{mathtools,etoolbox}
\usepackage[ruled, linesnumbered]{algorithm2e}
\usepackage[none]{hyphenat}
\usepackage[utf8]{inputenc}
\usepackage[english]{babel}
\usepackage[T1]{fontenc}
\usepackage{breqn}

\usepackage{amsthm}
\usepackage{xfrac}
\usepackage{nicefrac}
\usepackage{booktabs}
\usepackage{flushend}
\usepackage{mathrsfs}
\DeclareMathAlphabet{\mathpzc}{OT1}{pzc}{m}{it}
\usepackage{multirow}
\usepackage{multicol}
\usepackage{bbm}
\usepackage{soul}
\usepackage{xfrac}
\usepackage{balance}
\usepackage{titlesec}
\usepackage[numbers]{natbib}
\newcommand{\sgn}{\text{sgn}}
\newcommand{\clamp}{\text{clamp}}

\hypersetup{colorlinks=true,urlcolor=blue,citecolor=red}
\usepackage[some, placement=top]{background}

\backgroundsetup{
        placement=top,
        position={8cm,1.5cm}, firstpage=true}

\SetBgScale{1}
\SetBgContents{\parbox{16cm}{\large \centering This paper has been accepted for publication at Robotics: Science and Systems (RSS), 2021.}}
\SetBgColor{gray}
\SetBgAngle{0}
\SetBgOpacity{0.9}

\begin{document}

\title{\LARGE \bf EVPropNet: Detecting Drones By Finding Propellers For Mid-Air Landing And Following}

\author{Nitin J. Sanket$^1$, Chahat Deep Singh$^1$, Chethan M. Parameshwara$^1$, Cornelia Ferm{\"u}ller$^1$,\\ Guido C.H.E. de Croon$^2$, Yiannis Aloimonos$^1$ 
\thanks{$^{1}$Perception and Robotics Group, University of Maryland Institute for Advanced Computer Studies, University of Maryland, College Park.}
\thanks{$^{2}$Micro Air Vehicle Laboratory, Delft University
of Technology.}
\thanks{\textit{Corresponding author: Nitin
J. Sanket.}}} 

\makeatletter
\g@addto@macro\@maketitle{
\begin{figure}[H]
  \setlength{\linewidth}{\textwidth}
  \setlength{\hsize}{\textwidth}
    \centering
    \includegraphics[width=\textwidth]{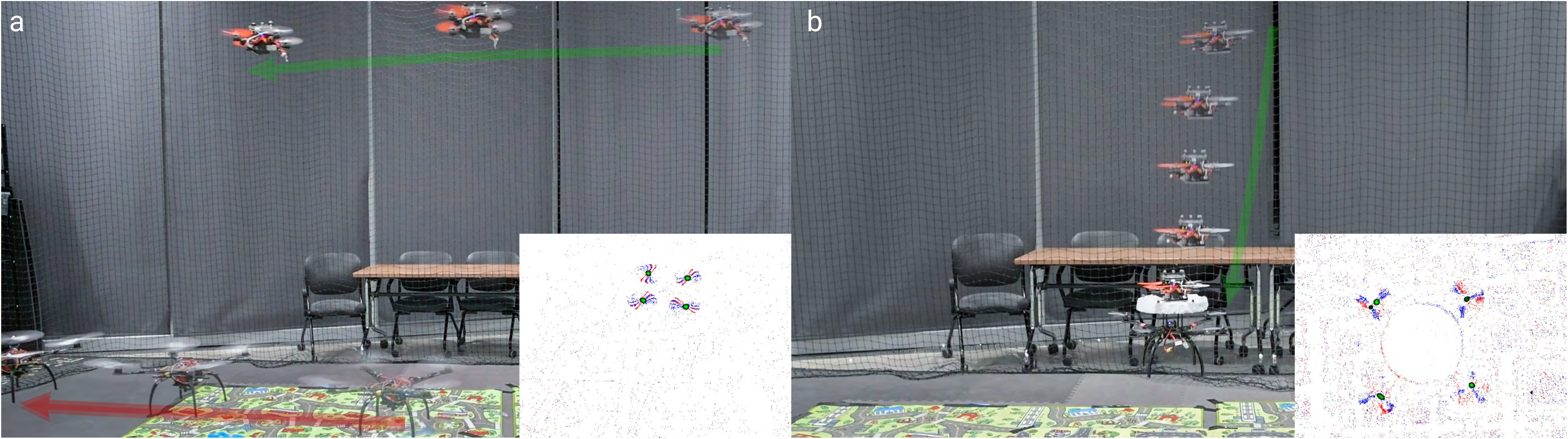}
    \caption{Applications presented in this work using the proposed propeller detection method for finding multi-rotors. (a) Tracking and following an unmarked quadrotor, (b) Landing/Docking on a flying quadrotor. Red and green arrows indicates the movement of the larger and smaller quadrotors respectively. Time progression is shown as quadrotor opacity. The insets show the event frames $\mathpzc{E}$ from the smaller quadrotor used for detecting the propellers of the bigger quadrotor using the proposed \textit{EVPropNet}. Red and blue color in the event frames indicate positive and negative events respectively. Green color indicates the network prediction. All the event images in this paper follow the same color scheme. Vicon estimates are shown in corresponding sub-figures of Fig. \ref{fig:Vicon}. \textit{All the images in this paper are best viewed in color on a computer screen at a zoom of 200\%.}}
    \vspace{-20pt}
    \label{fig:Banner}
    \end{figure}
}
\maketitle
\thispagestyle{plain}
\pagestyle{plain}

\setcounter{figure}{1}

\begin{abstract}
The rapid rise of accessibility of unmanned aerial vehicles or drones pose a threat to general security and confidentiality. Most of the commercially available or custom-built drones are multi-rotors and are comprised of multiple propellers. Since these propellers rotate at a high-speed, they are generally the fastest moving parts of an image and cannot be directly ``seen'' by a classical camera without severe motion blur. We utilize a class of sensors that are particularly suitable for such scenarios called event cameras, which have a high temporal resolution, low-latency, and high dynamic range.

In this paper, we model the geometry of a propeller and use it to generate simulated events which are used to train a deep neural network called \textit{EVPropNet} to detect propellers from the data of an event camera. \textit{EVPropNet} directly transfers to the real world without any fine-tuning or retraining. We present two applications of our network: (a) tracking and following an unmarked drone and (b) landing on a near-hover drone. We successfully evaluate and demonstrate the proposed approach in many real-world experiments with different propeller shapes and sizes. Our network can detect propellers at a rate of 85.1\% even when 60\% of the propeller is occluded and can run at upto 35Hz on a 2W power budget. To our knowledge, this is the first deep learning-based solution for detecting propellers (to detect drones). Finally, our applications also show an impressive success rate of 92\% and 90\% for the tracking and landing tasks respectively. 
\end{abstract}


\section*{Supplementary Material}
The accompanying video and code are available at \url{http://prg.cs.umd.edu/EVPropNet}. 

\section{Introduction}
Aerial robots or drones have become ubiquitous in the last decade due to their utility in various fields such as exploration \cite{Exploration, mcguire2017towards, mcguire2019minimal, sanket2020morpheyes}, inspection \cite{Inspection}, mapping \cite{Mapping}, search and rescue \cite{SearchAndRescue, GapFlyt}, and transport \cite{Mellinger2013}. The low-cost and wide availability of commercial drones for photography, agriculture and hobbying has skyrocketed drone sales \cite{DroneMarket}. This has also given rise to a series of malicious drones which threaten the general security and confidentiality. This necessitates the detection of drones. To make this problem hard, drones come in various shapes and sizes and generally do not carry any distinct visual  on them to make them easy for visual detection. To this end, we propose to detect an unmarked drone by detecting the most ubiquitous part of a drone -- the propeller. It is serendipitous that most of the common drones are multi-rotors and have more than one propeller, making their detection using the proposed approach easier. Detecting propellers is a daunting task for classical imaging cameras since it would require an extremely short shutter time and high sensitivity which make such sensors expensive and bulky. A class of sensors designed by drawing inspiration from nature that excel at the task of low-latency and high-temporal resolution data are called event cameras \cite{eventsurvey, dvspaper}. Recent advances in sensor technologies have increased the spatial resolution of these sensors by about 10$\times$ in the last 5 years \cite{samsungdvs}. These event cameras output per-pixel temporal intensity differences caused by relative motion with microsecond latency instead of traditional images frames. We utilize the fact that propellers are moving much faster than any other part of the scene. The problem formulation and our contributions are described next.

\subsection{Problem Formulation and Contributions}
An event camera (moving or stationary) is looking at a flying drone with at-least one spinning propeller and our goal is to locate the propeller's center on the image plane. A summary of our contributions are:
\begin{itemize}
    \item We simplify the geometric model of a propeller for the projection on the image plane which is used to generate event data.    
    \item A deep neural network called \textit{EVPropNet} trained on the simulated data which generalizes to the real-world without any fine-tuning or re-training for different propellers.
    \item Two specific applications of our \textit{EVPropNet}: (a) Tracking and following an unmarked moving quadrotor (Fig. \ref{fig:Banner}{\color{red}a}), (b) Landing on a near-hover quadrotor (Fig. \ref{fig:Banner}{\color{red}b}) evaluated with on-board computation and sensing.
    \item Finally, we make our network $\texttt{EdgeTPU}$ optimized so that it can run at 35Hz with a power budget of just 2W enabling deployment on a small drone.
\end{itemize}

\subsection{Related Work}
We subdivide the related work into three parts: detection of an unmarked drone based on appearance (on a classical camera), detection of a marked collaborative drone and detection of moving segments using event cameras. 

\subsubsection{Appearance based drone detection:} Classical RGB image based drone detection is an instance of object detection and has been accomplished by methods like Haar cascade detectors, with the newer deep learning based detectors such as YoLo \cite{yolov3} topping the accuracy charts \cite{pawelczyk2020real, unlu2019deep, schilling2020vision}. One can clearly see that these methods work well when the drone is large in the frame and against a bright sky, thereby detecting the contour of the drone from its silhouette. Pawe{\l}czyk \textit{et al.} \cite{pawelczyk2020real} show extensive results on how the state-of-the-art appearance based drone detectors fail when the drones are against a non-sky background (such as trees which are very common). 

\subsubsection{Marked collaborative drone detection:} 
Marked drones are detected using a set of fiducial markers either for leader-follower configurations \cite{walter2019uvdar}, swarming behaviors \cite{mateos2020apriltags} or for docking \cite{li2019modquad}. Most commonly, a visual fiducial marker based on April Tags \cite{krogius2019flexible} or CC-Tags \cite{calvet2016Detection} is used for these tasks due to their robustness and near-invariance to angles. Moreover, they also provide the ability to distinguish between different tags which are particularly useful for tracking multiple drones. Li \textit{et al.} \cite{li2019modquad} designed a custom tag similar to the CC-Tag and showed that it can be used for precise control for docking. On the contrary, Walter \textit{et al.} \cite{walter2019uvdar} demonstrated the usage of Ultra-Violet (UV) spectrum which is robust to changing environmental conditions such as changing illumination and the presence of undesirable light sources and their reflection.

\subsubsection{Moving Object Segmentation Using Event Cameras:}
Event cameras, as described earlier are tailor made for detecting the parts of the image which have motion different to that of the camera (this task is commonly called motion segmentation). Mitrokhin \textit{et al.} \cite{mitrokhin2018event} developed one of the first motion segmentation frameworks using event cameras for challenging lighting scenes highlighting the efficacy of event cameras to work at high-dynamic range scenes for fast moving objects. Stoffregen \textit{et al.} \cite{stoffregen2019event} introduced an Expectation-Maximization scheme for segmenting the motion of the scene into various parts which was further improved in-terms of speed and accuracy by Parameshwara \textit{et al.} \cite{parameshwara0} by proposing a motion propagation method based on cluster keyslices. The concept of motion segmentation has also been deployed on quadrotors for detection of other moving objects (including other unmarked drones) with a monocular \cite{sanket2020evdodgenet} and a stereo event camera \cite{falanga2020dynamic}. 



\subsection{Organization of the paper}
We first describe a geometric model of the propeller and then derive a simplified model of it's image projection in Sec. \ref{sec:PropModel}. The geometric model is then used generate event data to train the proposed \textit{EVPropNet} as described in Sec. \ref{sec:EVPropNet}. We then present two applications of our network: (a) Tracking and following an unmarked drone and (b) Landing on a near-hover drone in Sec. \ref{sec:Applications}. We then present extensive quantitative evaluation of our network and applications along with qualitative results on different real world propellers in Sec \ref{sec:Expts}.  Finally, we conclude the paper in Sec. \ref{sec:Conc} with parting thoughts on future work.

\begin{figure*}[t!]
    \centering
    \includegraphics[width=\textwidth]{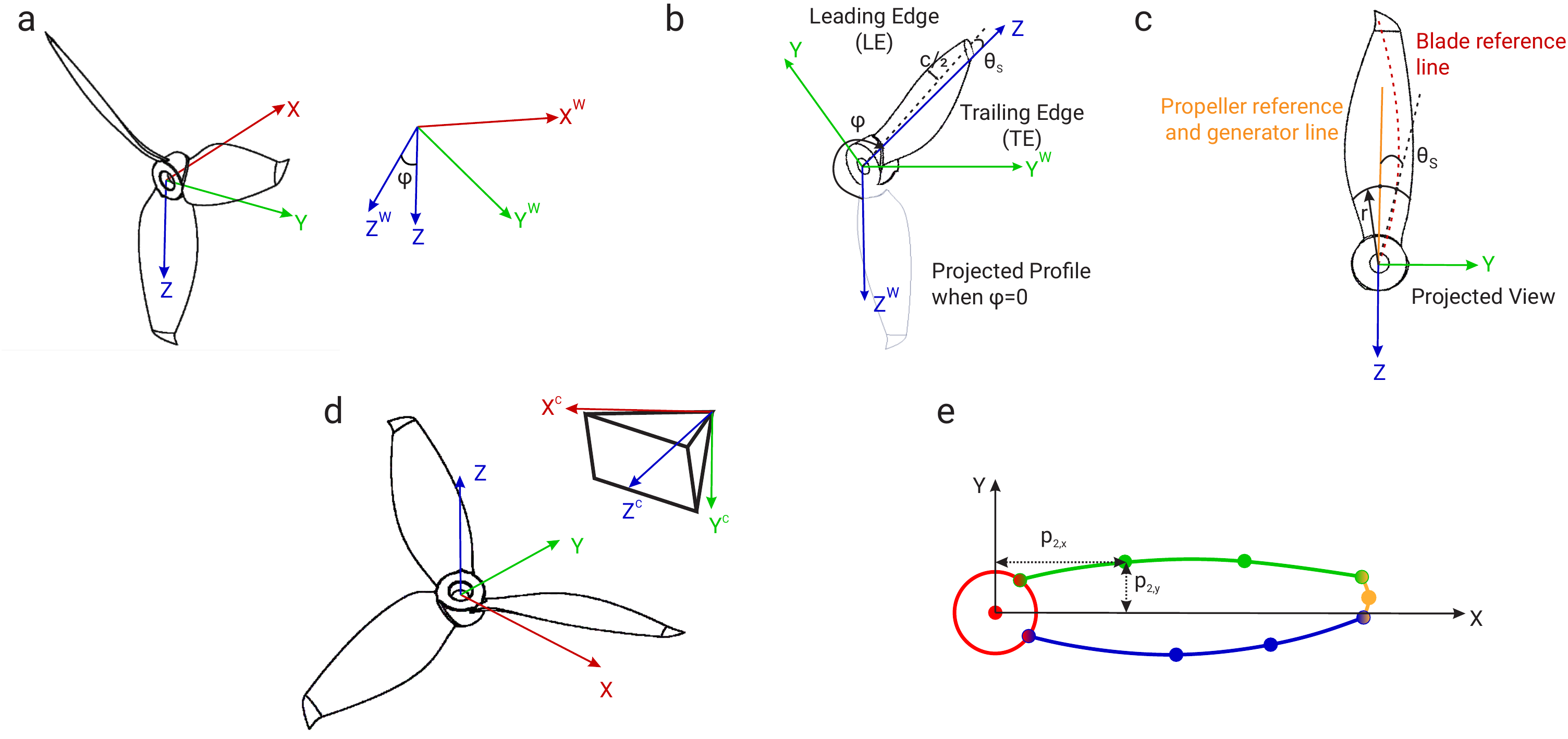}
    \caption{(a) Coordinate frames used for the geometric modelling of a propeller, (b) Blade coordinate definition, (c) Skew definition, (d) Coordinate axes for propeller projection on camera, and (e) Simplified model of the projection of the propeller blade; Each color represents a single spline and points with same color denote knots used to fit the cubic spline. Bi-color points are used as knots for both the splines of respective color. See Table \ref{tab:PropParams} for a tabulation of the variables used in this figure.}
    \label{fig:CoordinateFrames}
\end{figure*}

\section{Geometric Modelling of a Propeller}
\label{sec:PropModel}
We first discuss a geometric model of a propeller \cite{carlton2018marine} and then describe how it's projection can be approximated with a set of cubic basis splines. A propeller's spine is constructed by rotating a straight line on a helicoidal surface. The coordinates of a point $\mathbf{x}$ on a surface formed by a straight line rotating about the $X$ axis and concurrently moving along this axis is given by

\begin{equation}
\mathbf{x} = \begin{bmatrix} \dfrac{p \phi}{2\pi} & r \sin \phi & r \cos \phi \end{bmatrix}^T
\end{equation}

Where $p$ is the pitch of the propeller, $r$ is the radius and $\phi$ is the angle of rotation in $YZ$ plane of the radius arm relative to the $Z^W$ axis (Fig. \ref{fig:CoordinateFrames}{\color{red}a}, also see Table \ref{tab:PropParams} for a tabulation of the parameters used in this derivation). Note that, $p$ here refers to the nose-tail pitch as it is the most common definition used by manufacturers. Now, the locus of the mid-chord points of a rotating right handed propeller blade with $\phi=0$ initially is given by 

\begin{equation}
\mathbf{x}_{c/2} = \begin{bmatrix}
-\left( i_G + \dfrac{p\theta_S}{2\pi}\right)  & -r\sin \left( \phi -\theta_S\right) & r\cos \left( \phi -\theta_S\right)
\end{bmatrix}^T
\label{eq:midpointlocus}
\end{equation}

Here, $\theta_S$ (Figs. \ref{fig:CoordinateFrames}{\color{red}b} and \ref{fig:CoordinateFrames}{\color{red}c}) denotes the skew angle and is defined as the angle between the line normal to
the shaft axis (called \textit{directrix} or \textit{propeller reference line}) and the line drawn through the shaft center line and the mid-chord point on the projected image of the propeller looking normally along the shaft center line and $i_G$ denotes the generator line rake (distance that is parallel to the
$X$-axis, from the directrix to the point where the helix
of the section at radius $r$ cuts the $X-Z$ plane). Extending Eq. \ref{eq:midpointlocus} for the leading and trailing edges of the blade (Fig. \ref{fig:CoordinateFrames}{\color{red}b}) gives us

\begin{align}
\mathbf{x}_{LE/TE} &=
\begin{bmatrix}
-\left( i_G + \dfrac{p\theta_S}{2\pi} + \dfrac{c}{2}\sin \theta\right)  \\ -r\sin \left( \phi -\theta_S \pm \dfrac{90c\cos \theta}{\pi r} \right) \\ r\cos \left( \phi -\theta_S \pm \dfrac{90c\cos \theta}{\pi r} \right)
\end{bmatrix}
\end{align}

Here $\theta=\tan^{-1}\left( \frac{p}{2\pi r}\right)$ is the pitch angle, $c$ is the chord length at a certain radius and $\pi$ and $\theta_S$ are in $^\circ$. 

The above set of equations define how the propeller's \textit{nose-tail line} can be generated in 3D. However, the blade section geometry is an aerofoil with a top and a bottom surface. A point on the top and bottom surfaces are given by

\begin{table}[t!]
\centering
\caption{Parameters used in geometric model of the propeller.}
\resizebox{0.7\columnwidth}{!}{
\label{tab:PropParams}
\begin{tabular}{cc}
\toprule
Parameter & Parameter  \\
Notation & Description \\
\hline
$p$ & Propeller Pitch (nose-tail) \\
$r$ & Radius \\
\multirow{2}{*}{$\phi$} & Angle of rotation of radius arm \\
& relative to $Z^W$ in $YZ$ plane \\
$i_G$ & Generator line rake \\
$\theta_S$ & Skew angle \\
$c$ & Chord length \\
$\psi$ & Chamber line slope at $x_c$\\
Subscripts $T$ and $B$ & Top and Bottom surfaces of blade\\
Subscripts $LE$ and $TE$ & Top and Bottom edges of blade\\
$K$ & Camera intrinsic matrix \\
$f$ & Camera focal length\\
$c_x$ and $c_y$ & Camera principal points\\
$\mathbf{p_i}$ & Spline control points\\
$N_{i,k}(t)$ & Spline basis function\\
\bottomrule
\end{tabular}}
\end{table}

\begin{equation}
\mathbf{x}_{T/B} = \begin{bmatrix}
x_c \mp y_t\sin \psi & y_c \pm y_t\cos \psi
\end{bmatrix}^T
\label{eq:TopBot}
\end{equation}

where $y_c$ is the $y$ offset from the chord line,  $y_t$ is the ordinate of the point in question and $\psi$ is the slope of the chamber line at the non-dimensional chordal position $x_c$. Now, if we consider the chord's mid point as the local origin, then a point's coordinates $\mathbf{x}$ are given by 

\begin{equation}
  \mathbf{x} =   \begin{bmatrix}
    -\left( i_G + \dfrac{p\theta_S}{2\pi}  \right) 
 \left(0.5c -x_c\right)\sin \theta + y_{u,B}\cos \theta \\
 r\sin\left( \theta_S - \dfrac{180\left( \left(0.5c - x_c\right)\cos \theta - y_{u,B}\sin \theta \right)}{\pi r} \right) \\
 r\cos\left( \theta_S - \dfrac{180\left( \left(0.5c - x_c\right)\cos \theta - y_{u,B}\sin \theta \right)}{\pi r} \right)
 \end{bmatrix}
\end{equation}

Where $y_u = y_c \pm y_t \cos \psi$ (Eq. \ref{eq:TopBot}). To convert $x$ into global coordinates $\mathbf{x}^W$, 

\begin{equation}
    \mathbf{x}^W = \begin{bmatrix}
    1 & 0 & 0\\
    0 & \cos \phi & -\sin \phi\\
    0 &\sin \phi & \cos \phi \\
    \end{bmatrix} \mathbf{x}
    \label{eq:propworld}
\end{equation}

where $\phi$ is the angle between two adjacent blades. Although the propeller thickness varies along its chord line, we can neglect this value since we are concerned with the projection of the propeller on the image plane assuming that distance from the camera is $\gg$ propeller thickness. We want to simulate how a propeller would ``look'' when imaged from a camera (which would be later converted into an event stream). We assume that the image is captured from a calibrated camera formulated using the pinhole model given by

\begin{equation}
    \mathbf{x} = K
    \begin{bmatrix}
    R, & T
    \end{bmatrix} \mathbf{X}; \quad K = \begin{bmatrix}
    f & 0 & c_x\\
    0 & f & c_y\\
    0 & 0 & 1
    \end{bmatrix}
    \label{eq:pinhole}
\end{equation} 

where $K$ is the camera calibration matrix, $f$ is the focal length and $c_x, c_y$ denotes the principle point, $\begin{bmatrix} R, & T \end{bmatrix}$ denotes the pose of the camera, $\mathbf{X}$ is the world point being imaged and $\mathbf{x}$ is the location of the point on the image plane (with reuse of variables). $\mathbf{X}$ in Eq. \ref{eq:pinhole} (see Fig. \ref{fig:CoordinateFrames}{\color{red}d} for definition of coordinate frames) is given by $\mathbf{x}^W$ from Eq. \ref{eq:propworld}. Although, this method is the most generative way to model a propeller, i.e., generate 3D points of the propeller and then project onto the image plane, it would be computationally very expensive for a high fidelity image, hence we approximate the projection of a propeller blade with a set of cubic basis splines \cite{ding2019efficient, weisstein} described next. Let the $n+1$ control points be $\mathbf{p_0}, \cdots, \mathbf{p}_n$ and $m+1$ knot vectors be $\{t_0, \cdots , t_m \}$, the spline curve $s\left(t\right)$ of degree $k$ is given by 

\begin{equation}
    s\left(t\right) = \sum_{i=0}^n \mathbf{p}_iN_{i,k}(t)
\end{equation}

Here, $N_{i,k}(t)$ is the basis function of degree $k$ and is computed recursively as 

\begin{align}
   N_{i,0}\left(t\right) &= \begin{cases}
      1 & \text{if } t_i\le t \le t_{i+1}\\
      0 & \text{otherwise}\\
    \end{cases}\\
    N_{i,k} \left(t\right) &= \dfrac{t-t_i}{t_{i+k}-t_i}N_{i,k-1}\left(t\right) + \dfrac{t_{i+k+1}-t}{t_{i+k+1}-t_{i+1}}N_{i+1,k-1}\left(t\right)
\end{align}

In particular, $m=n+k+1$ and we utilize the uniform B-spline, i.e., all the knots are uniformly distributed and are evaluated using the procedure described in \cite{qin2000general}. We model each propeller blade using 4 cubic B-splines: one for the hub, one for the top part of the blade, one for the bottom part of the blade and one for the tip of the blade (Fig. \ref{fig:CoordinateFrames}{\color{red}e}). Each blade is replicated at a uniform angular spacing for the required number of blades (i.e., for a 3 bladed propeller, the blades would have an angle of 120$^\circ$ between them).



\section{EVPropNet}
\label{sec:EVPropNet}
We will now discuss how the geometric model of the propeller is used to generate event data. Then we describe the network architecture and loss function used to train \textit{EVPropNet}.
\subsection{Event Generation}
As explained earlier, we now have a single image of a propeller with the required number of blades at the required high resolution. We overlay this propeller image on top of a random real image background from the MS-COCO dataset \cite{MSCOCO} at a random starting angle $\theta_{HB}$ (angle the propeller reference line makes with the propeller $Y$ axis), we denote this as image $\mathcal{I}_t$. We then perform the same procedure for a $\theta_{HB} + \delta\theta$ angle (with the same background) to generate the image $\mathcal{I}_{t+\delta t}$. Here, $\delta \theta = \omega\delta t$ is the angle the blade would rotate depending on the rotational speed of the propeller $\omega$ and the event frame integration time $\delta t$. We use a simple model for the event camera and events are triggered at a location $\mathbf{x}$ when
\begin{equation}
  \Vert \log \left(\mathcal{I}_t\left(\mathbf{x}\right) \right)- \log \left(\mathcal{I}_{t+\delta t}\left(\mathbf{x}\right) \right)\Vert_1 \ge \tau
\end{equation}

The event stream/cloud $\mathcal{E}$ is represented by 
\begin{equation}
   \mathcal{E} = \{  \begin{bmatrix}
     \mathbf{x} & t & \sgn\left( \log \left(\mathcal{I}_t\left(\mathbf{x}\right) \right)- \log \left(\mathcal{I}_{t+\delta t}\left(\mathbf{x}\right) \right) \right) 
    \end{bmatrix}^T \}
\end{equation}

\begin{figure}[t!]
    \centering
    \includegraphics[width=\columnwidth]{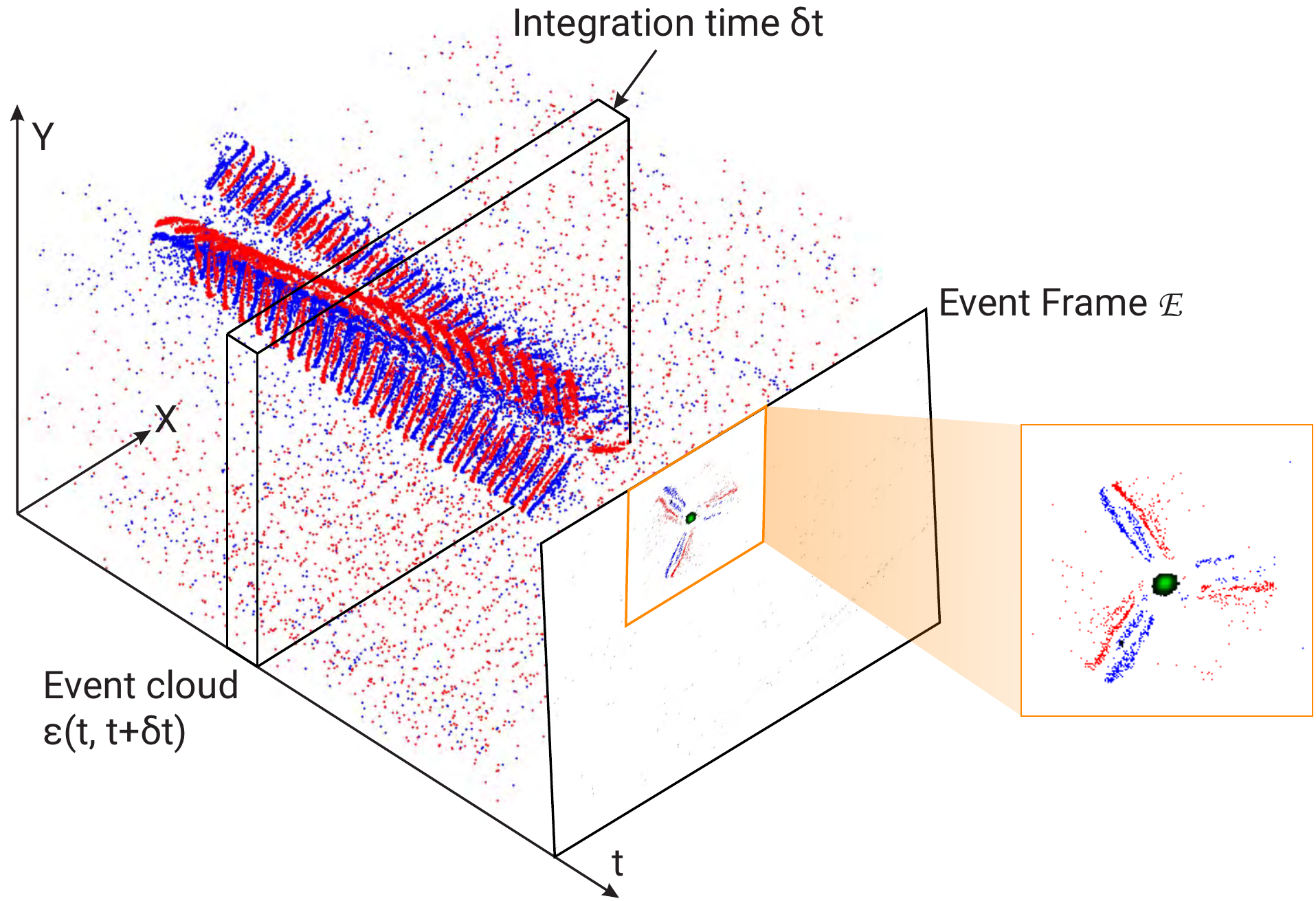}
    \caption{Spatio-temporal event cloud $\mathcal{E}$ and Event frame $\mathpzc{E}$. The cloud shows that the propeller creates a helix in the spatio-temporal domain. The zoomed in view shows the propeller with positive events colored red and negative events colored blue along with network prediction as green with the color saturation indicating confidence.}
    \label{fig:EventCloud}
\end{figure}

Where $\tau$ is a user defined threshold and $\mathbf{x}$ is the pixel location. $\mathcal{E}$ is called the event cloud which is used to create the so-called \textit{Event-frame} $\mathpzc{E}$ (Fig. \ref{fig:EventCloud} shows how $\mathcal{E}$ and $\mathpzc{E}$ look) which is used as the input to the network.
\begin{equation}
    \mathpzc{E} = \sgn\left(\mathbb{E}_t\left( \text{Pol}\left(\mathcal{E} \left( t, t+\delta t\right) \right)\right)\right)
\end{equation}
Here, $\mathbb{E}_t$ denotes the averaging operator only in time axis, Pol denotes the polarity values are extracted per pixel (last row of each element of $\mathcal{E}$).

\begin{figure}[t!]
    \centering
    \includegraphics[width=\columnwidth]{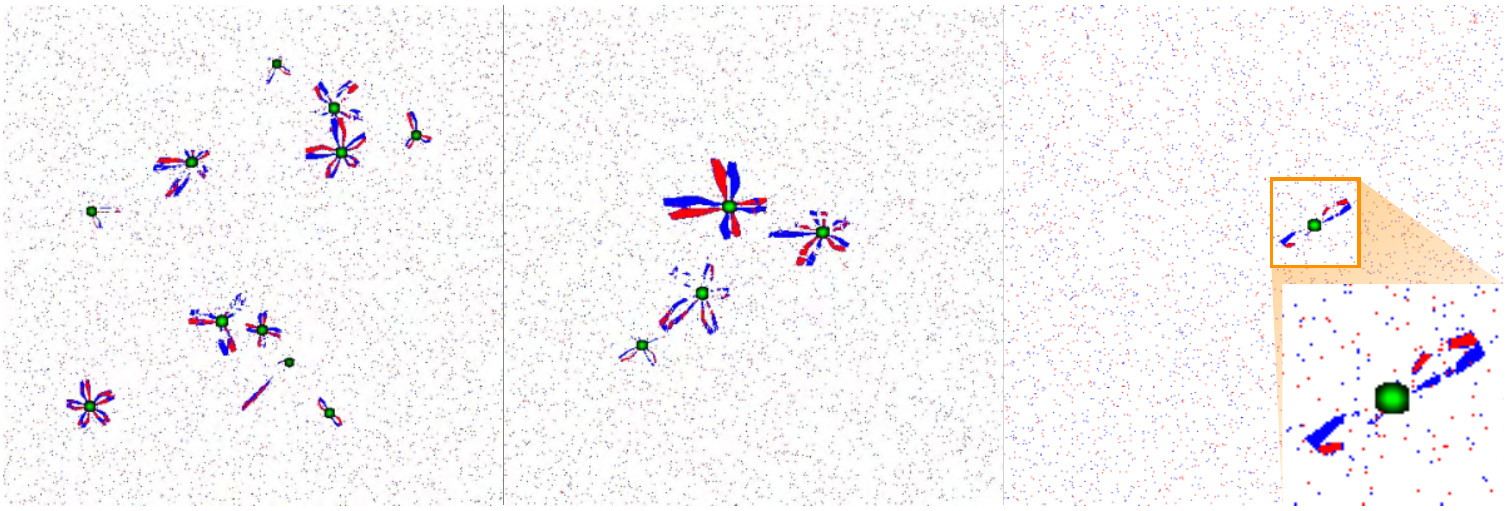}
    \caption{Sample event images $\mathpzc{E}$ from the generated synthetic dataset used to train \textit{EVPropNet}. Here  red  and  blue  colors  show  positive  and  negative  events  respectively. Green  color  indicates  our  ground truth label with the color saturation indicating confidence as defined by Eq. \ref{eq:Gaussian}.}
    \label{fig:DatasetImg}
\end{figure}

\begin{figure*}[t!]
    \centering
    \includegraphics[width=\textwidth]{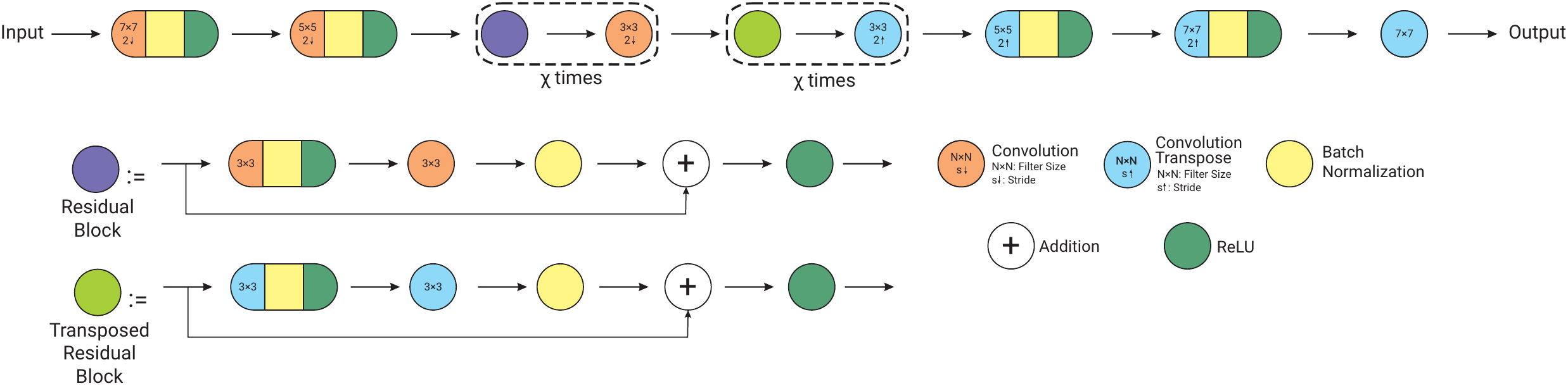}
    \caption{Network architecture for \textit{EVPropNet} ($\chi$ is a hyperparameter along with expansion rate -- rate at which the number of neurons grow after each block). If no down/up-sampling rate is shown, it is taken to be 1. \textit{This image is best viewed on the computer screen at a zoom of 200\%}.}
    \label{fig:NetworkArchitecture}
\end{figure*}

\subsection{Data Generation}
We generate 10K event frames $\mathpzc{E}$ for training our network (See Fig. \ref{fig:DatasetImg} for sample images with labels overlaid). Each event frame contains upto $N$ propellers (set to 12 in our case) with number of blades per propeller randomized from 2 to 6. The events for each propeller are obtained by varying $\tau$ as a gaussian random variable to provide some randomness in data generation along with randomization of the color of the propeller in $\mathcal{I}_t$ (same color is used for $\mathcal{I}_{t+\delta t}$) along with varying $\omega$ (rotational speed of the propeller, this is equivalent to varying the integration time $\delta t$). We also vary the background image for every propeller from the MS-COCO dataset. Each propeller is also warped using a random homography matrix to account for different camera angles along with scaling them (setting the pixel size of the propeller in the event image) to account for distance variation from the camera. Finally, we also vary the shape of each propeller by varying the basis spline parameters (to include bullnose and normal type propellers as well). See Fig. \ref{fig:DatasetImg} for some sample images from the dataset used to train \textit{EVPropNet}. Note that, we do not use an event simulator like ESIM\cite{ESIM} to generate events since we only require $\mathcal{I}_t$ and $\mathcal{I}_{t+\delta t}$ which are directly constructed, hence this process is multiple orders of magnitude faster than real-time and parallelized. 

\subsection{Network Architecture and Loss Function}
We choose an encoder-decoder architecture based on the ResNet \cite{he2016deep} backbone (Fig. \ref{fig:NetworkArchitecture}) as it has the best accuracy and speed tradeoff \cite{bianco2018benchmark, sanket2020prgflow} with 2.7M parameters and 10MB model size. We train our network using simple mean square loss $\mathcal{L} = \mathbb{E}\left(\left( \hat{p} - \tilde{p} \right)^2\right)$ between the ground truth $\hat{p}$ and prediction $\tilde{p}$. $\hat{p}$ is obtained by Gaussian smoothing the perfect label $\hat{p_0}$ (binary mask) as given by Eq. \ref{eq:Gaussian} ($\sigma$ is the variance)  to account for small distortion introduced by approximation of propeller shape. This approach is similar to the one introduced in  \cite{foehn2020alphapilot}.

\begin{equation}
\hat{p} ={\frac {1}{2\pi \sigma ^{2}}}e^{-{\left(\Vert \hat{p_0}\Vert_2 / 2\sigma ^{2}\right)}}
\label{eq:Gaussian}
\end{equation}


We choose the number of residual and transposed residual blocks $\chi$ as 2 and expansion factor as 2 (factor with which number of neurons grow after every block in Fig. \ref{fig:NetworkArchitecture}).

\begin{figure}[t!]
    \centering
    \includegraphics[width=\columnwidth]{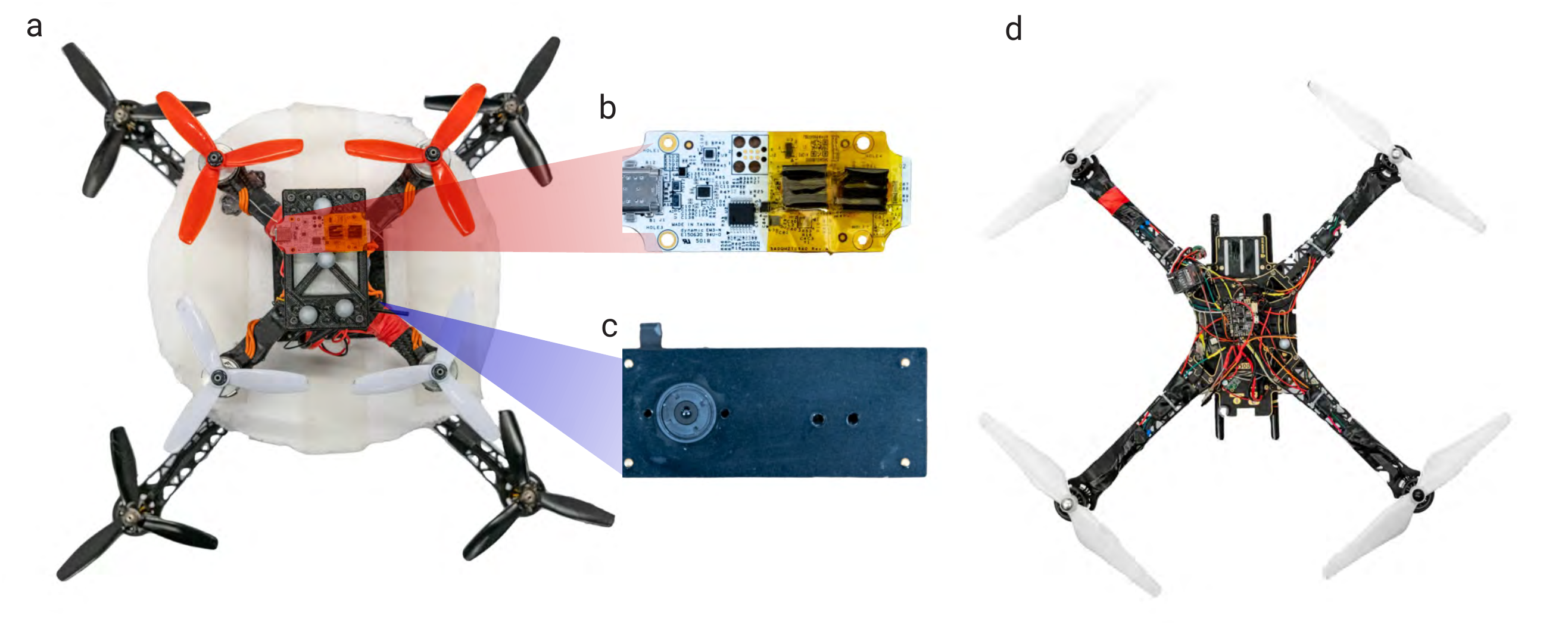}
    \caption{(a) Smaller quadrotor on the bigger quadrotor used for landing experiments (Sec. \ref{subsec:Following}), (b) Gutted Coral USB Accelerator with custom heat sink used to run the neural networks, (c) Samsung Gen 3 DVS sensor used for experiments, (d) Bigger quadrotor used in the following experiments (Sec. \ref{subsec:Landing}).}
    \label{fig:Quads}
\end{figure}

Finally, \textit{EVPropNet} was trained with a learning rate of 1e-4 using ADAM optimizer with a batch size of 32 for 50 epochs.

\section{Applications}
\label{sec:Applications}
We describe two applications of our propeller detection, i.e., following an unmarked moving quadrotor and landing on a near-hover quadrotor. 

\subsection{Following}
\label{subsec:Following}
In this application, the goal is to track and follow a quadrotor (or a multirotor in general) either for swarming or reconnaissance purposes. We detect the quadrotor as the centroid of filtered propeller detections as described in Sec. \ref{subsec:Filtering}. The control policy for altitude is to maintain the area of the quadrilateral joining the propeller points constant and the control policy for roll and pitch is to maintain the centroid of the quadrilateral in the center of the image and are given by:
\begin{align}
\mathbf{u}_\phi(t) &= K_{p, \phi} e_x(t) + K_{i, \phi}\int_0^\tau e_x(\tau)d\tau +  K_{d, \phi}\frac{de_x(t)}{dt}\\
\mathbf{u}_\theta(t) &= K_{p, \theta} e_y(t) + K_{i, \theta}\int_0^\tau e_y(\tau)d\tau +  K_{d, \theta}\frac{de_y(t)}{dt}\\
\mathbf{u}_T(t) &= K_{p, T} e_A(t) + K_{i, T}\int_0^\tau e_A(\tau)d\tau +  K_{d, T}\frac{de_A(t)}{dt}
\end{align}

Where $e_x$ and $e_y$ denote the difference between detected quadrilateral center on the image plane and the center of the image and $e_A$ denotes the difference between area to maintain and current area.

\subsection{Landing}
\label{subsec:Landing}
For the second application, the goal is to land on a near-hover quadrotor either for in-air battery switching or infiltration of a hostile drone. We utilize the following key observations from \cite{jain2020flying}:
\begin{itemize}
    \item The quadrotor flying above experiences negligible aerodynamic disturbances from mutual interaction. 
    \item Forces in direction normal to the downwash are negligible and those in the downwash direction are significant.
    \item The aerodynamic torques disturb the bottom quadrotor so that it aligns vertically with the top quadrotor.
\end{itemize}
The smaller quadrotor (which will land) explores the area for any other quadrotor (or any multirotor in general), once it detects a multirotor, it switches to the align maneuver, where we perform the following control policy for roll $\phi$ and pitch $\theta$ axes ($X$ and $Y$ respectively) for aligning the center of the detected quadrotor (centroid of filtered propeller detections as described in Sec. \ref{subsec:Filtering}) with the center of the image:
\begin{align}
\mathbf{u}_\phi(t) &= K_{p, \phi} e_x(t) + K_{i, \phi}\int_0^\tau e_x(\tau)d\tau +  K_{d, \phi}\frac{de_x(t)}{dt}\\
\mathbf{u}_\theta(t) &= K_{p, \theta} e_y(t) + K_{i, \theta}\int_0^\tau e_y(\tau)d\tau +  K_{d, \theta}\frac{de_y(t)}{dt}
\end{align}

Where $e_x$ and $e_y$ denote the difference distance between detected quadrotor center on the image plane and the center of the image. Once the errors $e_x$ and $e_y$ are lower than a threshold, we decrease altitude at a constant rate, checking for $x$ and $y$ alignment at every control loop and re-aligning as necessary. Once we are close to the big quadrotor (on which the smaller quadrotor will land), we initiate the land command. 

\subsection{Quadrotor Location from Detected Propellers and Filtering}
\label{subsec:Filtering}
We filter each propeller location on the image plane using a linear Kalman filter \cite{kalman1960new}. The motion model is a constant optical flow model. Once we obtain detections with a confidence above a certain threshold, the filtered propeller locations are used to compute the centroid of the quadrilateral (for the quadrotor case, polygon in general) which is used for control (to compute $e_x$ and $e_y$), along with the area for altitude control. 

\section{Experimental Results and Discussion}
\label{sec:Expts}
\subsection{Quadrotor Setup}
All our experiments are performed with quadrotors for their minimal hardware complexity and cost-effectiveness but they can directly be adapted to any multirotor vehicle. Our smaller quadrotor is a custom built platform on a QAV-X 210mm sized (motor center to motor center diagonal distance) racing frame. The motors used are T-Motor F40III KV2400 mated to 5040$\times$3 propellers (Fig. \ref{fig:Quads}{\color{red}a}). The lower level controller and position hold is handled by ArduCopter 4.0.6 firmware running on the Holybro Kakute F7 flight controller mated to an optical flow sensor and TFMini LIDAR as altimeter source. All the higher level navigational commands are sent by the companion computer (NVIDIA Jetson TX2 running Linux for Tegra$^{\text{\textregistered}}$) using RC-Override to the flight controller running in Loiter mode using MAVROS. The event camera used is a Samsung Gen-3 Dynamic Vision Sensor \cite{samsungdvs} with a resolution of 640$\times$480 px. (Fig. \ref{fig:Quads}{\color{red}c}). and is mounted facing forward tilted down by 45$^\circ$ for the following experiment and facing down for the landing experiment. \textit{All the computations and sensing are done on-board with no use of an external motion capture system.} Our neural network runs on a gutted Google Coral USB Accelerator with a custom heatsink attached to the TX2. The quadrotor take-off weight including the battery is 680 g and has a thrust to weight ratio of 5:1. Our network runs at 35Hz on the Coral accelerator (See Fig. \ref{fig:Quads}{\color{red}b}. Implementation details are given in Sec. \ref{subsec:Impl}) and our planning and control algorithms run at 15 Hz on the TX2. 

The larger quadrotor used in the following experiment is built on a S500 frame with DJI F2312 960KV motors mated to white colored 9450$\times$2 propellers (Fig. \ref{fig:Quads}{\color{red}d}). Same avionics components are used as the smaller quadrotor. 

The larger quadrotor used in the landing experiment is built on a S500 frame with T-Motor F80 Pro KV2500 motors mated to black colored 6040$\times$3 propellers (Fig. \ref{fig:Quads}{\color{red}a}). Same avionics components are used as the smaller quadrotor and ArduCopter firmware holds the position in Loiter mode during experiments with all the sensor fusion, control and planning handled by the flight controller. \textit{The area where the smaller quadrotor can land is of radius 135mm, which gives a tolerance of just 30mm on each side.}

\subsection{Experimental Results And Observations}
\subsubsection{Quantitative Evaluation of \textit{EVPropNet}}
In the first case study we discuss quantitative evaluation results  of our propeller detection results for varying resolution of propeller blade $r_{\text{px}}$ (the bounding box size of the propeller would be $2r_{\text{px}}$ and is directly correlated with real-world propeller size $r$), number of blades $N_{\text{blades}}$, noise probability $p_n$, data miss probability $p_b$, different camera roll and pitch angles ($\phi$ and $\theta$ respectively). Formally, $p_n$ denotes the probability with which a pixel can have error (equally likely to be either a positive or negative event) and $p_b$ denotes the probability with which the pixel where the propeller data exists did not fire either due to a dead-pixel or camouflage with the background. We use the following metric to denote a successful detection of a propeller.

\begin{equation}
    \text{Success} := \nicefrac{\mathcal{G}\cap\mathcal{D}}{\mathcal{G}\cup\mathcal{D}} \ge 0.5; \mathcal{G}: \text{Ground Truth, }\mathcal{D}: \text{Detection}
\end{equation}

Detection Rate DR is given by  $\text{DR}=\mathbb{E}\left(\text{Success}\right)$, where $\mathbb{E}$ is the expectation/averaging operator. The results are presented in Table \ref{tab:DRProp}. When not specified, the values for the parameters are given as follows: $r_{\text{px}}=\{20,30,40,50,60\}, N_{\text{px}}=\{2,3,4,5,6\}, \text{RPM}=\{\text{5K,10K,20K,30K,40K}\}, p_n=\{0, 0.01, 0.02\}, p_b=\{0,0.15,0.3, 0.45, 0.6\}, \phi=\{0^\circ,10^\circ,20^\circ,30^\circ,60^\circ\}$ and $\theta=\{0^\circ,10^\circ,20^\circ,30^\circ,60^\circ\}$.

\begin{table*}[h!]
\centering
\caption{Detection Rate (\%) $\uparrow$ of \textit{EVPropNet} for variation in parameters.}
\resizebox{0.6\textwidth}{!}{
\label{tab:DRProp}
\begin{tabular}{lccccclccccc}
\toprule
\multirow{2}{*}{(a)} & \multicolumn{5}{c}{$r_{\text{px}}$ (px.) for $\phi=\theta=0^\circ$} & \multirow{2}{*}{(b)} & \multicolumn{5}{c}{$N_{\text{blades}}$ for $\phi=\theta=0^\circ$}  \\
 \cline{2-6} \cline{8-12}
& 20 & 30 & 40 & 50 & 60 & & 2 & 3 & 4 & 5 & 6\\
 \hline
& 78.9 & 90.4 &	94.4 &	97.6 & 93.9 & & 77.9 & 94.1 & 96.3 & 92.8 & 94.1 \\
\hline
\midrule
\multirow{2}{*}{(c)} & \multicolumn{5}{c}{RPM (min$^{-1}$) for $\phi=\theta=0^\circ$} & \multirow{2}{*}{(d)} & \multicolumn{5}{c}{$p_n$ for $\phi=\theta=0^\circ$} \\
 \cline{2-6} \cline{8-12}
 & 5K & 10K & 20K & 30K & 40K & & 0 & 0.01 & 0.02 & & \\
 \hline 
 & 71.5 & 91.7 & 97.1 & 98.1 & 96.8 & & 92.6 & 91.4 & 89.1 & & \\
 \hline
\midrule
\multirow{2}{*}{(e)} & \multicolumn{5}{c}{$p_b$ for $\phi=\theta=0^\circ$} & \multirow{2}{*}{(f)} & \multicolumn{5}{c}{$\phi$ ($^\circ$) for $p_n=p_b=0$} \\
\cline{2-6} \cline{8-12}
 & 0 & 0.15 & 0.3 & 0.45 & 0.6 & & 0 & 10 & 20 & 30 & 60 \\
\hline
& 97.3 & 94.1 & 95.5 & 88.8 & 79.5 & & 97.6 & 	94.7 &	94.1 & 94.1 & 89.1\\
 \hline
\midrule
 & & & \multirow{2}{*}{(g)} & \multicolumn{5}{c}{$\theta$ ($^\circ$) for $p_n=p_b=0$} & & &  \\
 \cline{5-9} 
 & & & & 0 & 10 & 20 & 30 & 60 & & &  \\
 \hline
  & & & & 97.6 & 96.5 & 93.4 & 93.6 & 86.7 & & & \\
 \hline
\bottomrule
\end{tabular}}
\begin{tiny}
\\

\end{tiny}
\end{table*}

We see from Table \ref{tab:DRProp}{\color{red}a} that DR increases with propeller size and then decreases, this is because as the amount of data increases, the results improve and but when the propeller is large ($r_{\text{px}}=60$), we observe an increase in false detections near the edges of the propeller blades, dropping the DR slightly (Table \ref{tab:DRProp}{\color{red}a}). A similar trend is observed with $N_{\text{blades}}$ and RPM with DR peaking for a 4 bladed propeller and at 10K RPM (Tables \ref{tab:DRProp}{\color{red}b} and \ref{tab:DRProp}{\color{red}c}). We also observe that with increase in $p_n$ (Table \ref{tab:DRProp}{\color{red}d}), the detection results are not affected significantly highlighting the robustness of our network. Even when 60\% of the propeller is camouflaged with the busy background, we obtain a DR of above 79\% (Table \ref{tab:DRProp}{\color{red}e}) with the DR decreasing with increase in camouflage amount as expected. From Tables \ref{tab:DRProp}{\color{red}f} and \ref{tab:DRProp}{\color{red}g}, we also observe that even with camera angles ($\phi$ and $\theta$) = $60^\circ$, we obtain a DR of above 85\%. Finally, we obtain an overall DR of 85.1\% for variations in all parameters and  90.9\% when no data is corrupted.

Also, if we define success for drone detection as detecting atleast $\eta$ propellers of the drone, we obtain the drone detection rate $\text{DR}_{\text{D}}$ as follows $\text{DR}_{\text{D}} = \left(1 - \left(1-\text{DR}\right)^\eta\right)$, where DR is the detection rate of a single propeller. For example, we would obtain a drone detection rate $\text{DR}_{\text{D}}$ of 97.7\%, 99.6\% and 99.9\% for a quadrotor, hexacopter and octocopter respectively even when only 50\% of the propellers are detected.

\subsubsection{Quantitative Evaluation of April Tags 3}
In the second case study, we evaluate how a custom designed passive fiducial marker would perform the task of detecting a drone (note that this is only applicable to a collaborative drone). In particular, we evaluate one of the most ubiquitous and robust passive fiducial markers April Tag 3 \cite{krogius2019flexible} (36h11 family) inspired from \cite{pfrommer2017penncosyvio}. The parameters are the same as the first case study. From Table \ref{tab:DRAT}, we observe that when the data is not missing (occluded or not correctly exposed), the April Tag detects the tags with an impressive DR (tag ID correctness is not considered) of 100\%, but the accuracy falls significantly to 61.9\% when data is missing (which is common in real-world due to high dynamic range scenarios and motion blur). It is also important to note the following reasons when a drone detection based on event camera based propeller detection will be better than a passive fiducial marker based detection. 

\begin{table}[t!]
\centering
\caption{Detection Rate (\%) $\uparrow$ of AprilTags 3 for amount of tag blocked.}
\resizebox{0.5\columnwidth}{!}{
\label{tab:DRAT}
\begin{tabular}{ccccc}
\toprule
 \multicolumn{5}{c}{$p_b$}  \\
\cline{1-5}
 0 & 0.15 & 0.3 & 0.45 & 0.6 \\
 \hline 
 100 & 91.5 & 73.3  & 40.5 & 4.0  \\
\hline
\bottomrule
\end{tabular}}
\end{table}




\begin{itemize}
    \item Detection of a non-collaborative drone for reconnaissance purposes 
    \item High dynamic range and adverse lighting scenarios including fast movement
    \item Area occupied by non-occluded propellers is generally $\gg$ area occupied by the fiducial marker in the center (Refer to Sec. \ref{subsec:Analysis} for a detailed analysis)
    \item When a major part of the fiducial marker would be generally occluded 
\end{itemize}

\subsubsection{Quantitative Evaluation of Appearance based drone detectors}
In the third case study, we compare drone detection using classical appearance based detectors from \cite{pawelczyk2020real}. We see that the Haar Cascade detectors have a DR of 55.2\% and the MobileNet deep learning based detector has a detection rate of 69.4\% which are far lower than those of the fiducial detector and our propeller detection method. 

\subsubsection{Performance Measure on different compute platforms}
In the fourth case study, we present speed and timing results for \textit{EVPropNet} on various commonly used computational platforms. We refer the readers to \cite{sanket2020prgflow} for a detailed description of the compute modules used in this case study. \textit{EVPropNet} has 2.7M parameters, a model size of $\sim$10MB and utilizes 17GOPs for a single forward pass. We can see from Table \ref{tab:DetSpeed} that running \textit{EVPropNet} to detect a drone by detecting propellers on the Google Coral Accelerator attached to the NVIDIA Jetson TX2 has the best speed and detection performance per unit power.



\begin{figure*}[t!]
    \centering
    \includegraphics[width=0.7\textwidth]{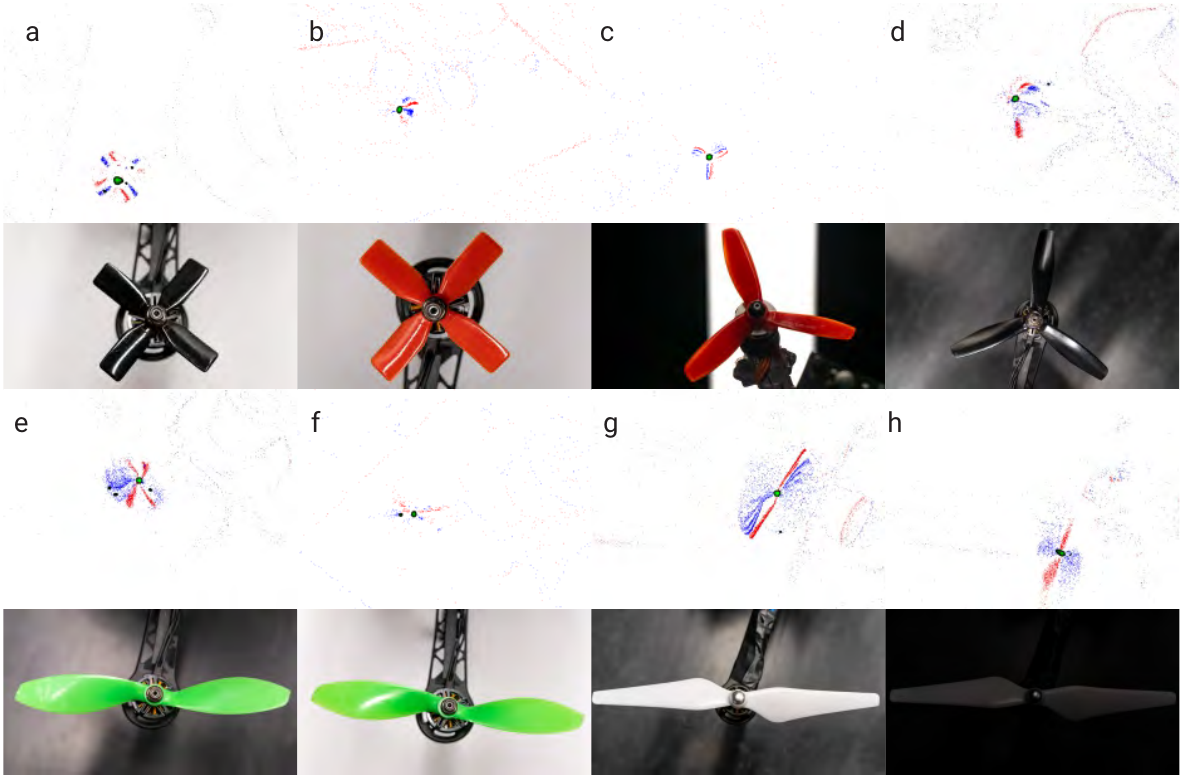}
    \caption{Top rows: Input event frame $\mathpzc{E}$ where red and blue colors show positive and negative events respectively. Green color indicates \textit{EVPropNet} prediction with the color saturation indicating confidence. Bottom rows: reference images of the propeller taken with a Nikon D850 DSLR (32dB dynamic range). Scenarios (a) to (h) are explained in Table \ref{tab:PropScenes}.}
    \label{fig:RealProps}
\end{figure*}

\begin{figure}[t!]
    \centering
    \includegraphics[width=\columnwidth]{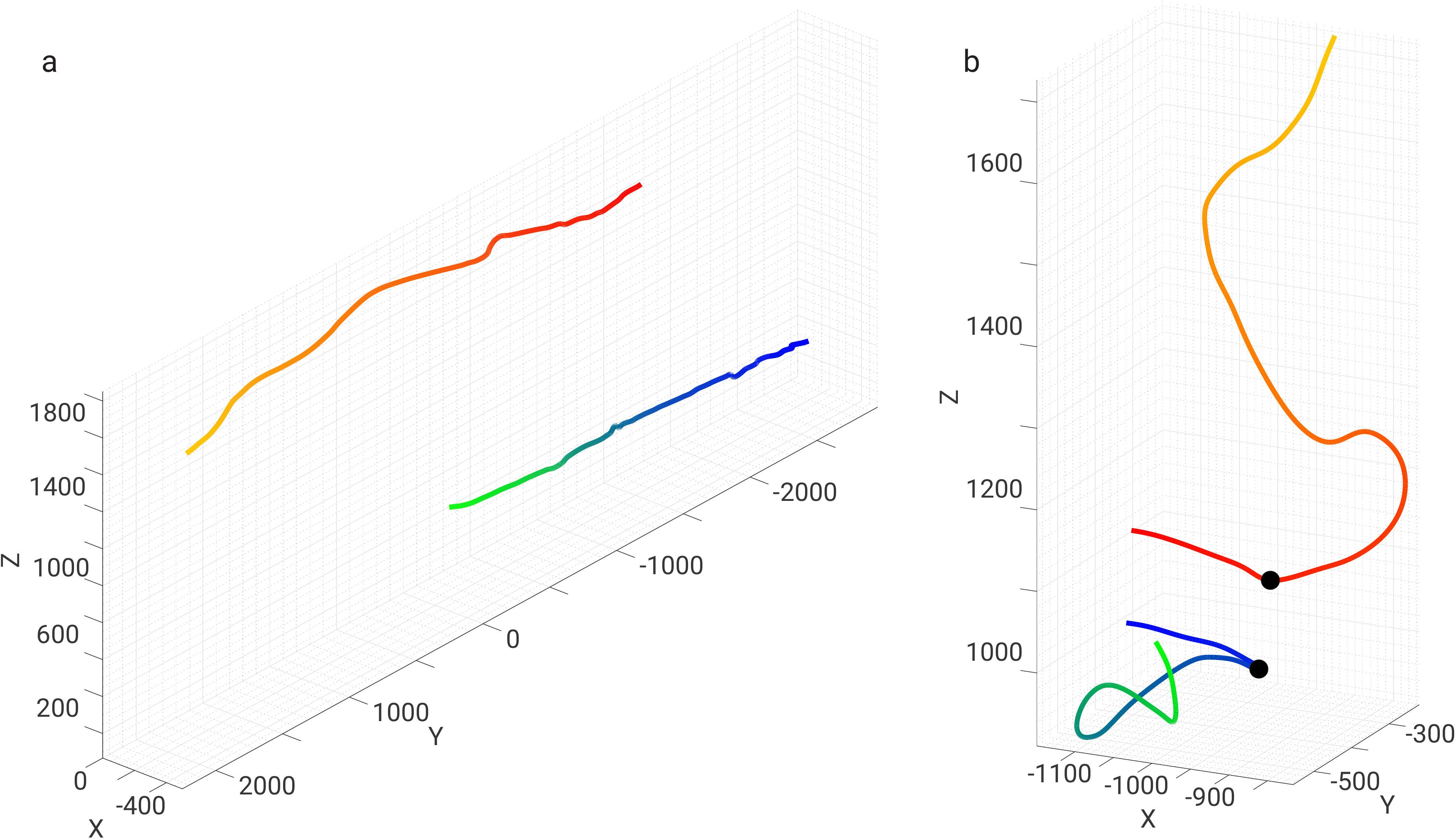}
    \caption{Vicon estimates for the trajectories of the smaller and larger quadrotor in the application experiments shown in Fig.  \ref{fig:Banner}. (a) Tracking and following, (b) Mid-air landing. Time progression is shown from yellow  to red for the smaller quadrotor and and green to blue for the bigger quadrotor. The black dots in (b) show the moment in time where the touchdown occured.}
    \label{fig:Vicon}
\end{figure}

\begin{table*}[h!]
\centering
\caption{Performance Metrics On Different Compute Modules.}
\resizebox{0.6\textwidth}{!}{
\label{tab:DetSpeed}
\begin{tabular}{cccccccc}
\toprule
\multirow{2}{*}{Method} & \multicolumn{5}{c}{(Ours)} & AprilTags 3 &  AprilTags 3 \\
& \multicolumn{5}{c}{\textit{EVPropNet} (Ours)} & (36h11) &  (16h5) \\
\hline
Computing  & PC  & PC &  \multirow{2}{*}{TX2$^\dagger$} &  \multirow{2}{*}{NCS2$^*$} &  \multirow{2}{*}{Coral$^*$} & \multirow{2}{*}{TX2} & \multirow{2}{*}{TX2}\\
Platform & (i9)  & (TitanXp) &  &  &  & &\\
\hline 
Speed $\uparrow$ & \multirow{2}{*}{8.6} & \multirow{2}{*}{133.4} & \multirow{2}{*}{10.5} & \multirow{2}{*}{4.5} & \multirow{2}{*}{35.2} & \multirow{2}{*}{7.0} & \multirow{2}{*}{41.3}\\
(Frames per second)   &  &   &   &   &  &   &  \\
\hline
Weight (g) $\downarrow$ &  -- & -- & 130 & 138 & 136 & 130 & 130\\
\hline
Peak Power (W) $\downarrow$ & 250 & 250 & 15 & 17 & 17 & 15 & 15\\
\hline
Speed/Unit $\uparrow$ & \multirow{2}{*}{0.03} & \multirow{2}{*}{0.53} & \multirow{2}{*}{0.7} & \multirow{2}{*}{0.27} & \multirow{2}{*}{2.07} & \multirow{2}{*}{0.47} & \multirow{2}{*}{2.75}\\
Power (FPS/W) &  &   &   &   &  &   &  \\
\hline
Detection Rate (\%) $\uparrow$ & 85.1 & 85.1 & 85.1 & 83.4 & 81.9 & 61.9 & 53.4 \\
\hline
Speed$\times$DR/Unit $\uparrow$& \multirow{2}{*}{2.55} & \multirow{2}{*}{45.10} & \multirow{2}{*}{59.57} & \multirow{2}{*}{22.52} & \multirow{2}{*}{\textbf{169.53}} & \multirow{2}{*}{29.09} & \multirow{2}{*}{146.85} \\
Power (FPS\%/W) &  &   &   &   &  &   &  \\
\bottomrule
\end{tabular}}
\begin{tiny}\\
$^\dagger$Active heatsink removed. $^*$Attached to TX2, outer casing removed and custom heatsink. 
\end{tiny}
\end{table*}

\begin{table}[htbp!]
\centering
\caption{Different Propeller Configurations Used for Qualitative Evaluation in Fig. \ref{fig:RealProps}.}
\resizebox{\columnwidth}{!}{
\label{tab:PropScenes}
\begin{tabular}{lcccccc}
\toprule
\multirow{2}{*}{Scenario} & Ref. & Prop. & Background & Prop. & Background Light & \multirow{2}{*}{$\cfrac{\text{Propeller Area}}{\text{Motor Area}}$} \\
 & Fig. & Color & Color & Radius (mm) & Intensity (lx) &  \\
\midrule
(a) & \ref{fig:RealProps}{\color{red}a} & Black & White & 50.8 & 240 & 2.3\\
(b) & \ref{fig:RealProps}{\color{red}b} & Red & White & 50.8 & 240 & 2.3\\
\multirow{2}{*}{(c)$^*$} & \multirow{2}{*}{\ref{fig:RealProps}{\color{red}c}} & \multirow{2}{*}{Red} & White and  & \multirow{2}{*}{63.5} & 564 and & \multirow{2}{*}{3.6} \\
 & & &  Black  & &  2 &\\
(d) & \ref{fig:RealProps}{\color{red}d} & Black & Black & 76.2 & 240 & 5.2\\
(e) & \ref{fig:RealProps}{\color{red}e} & Green & Black & 88.9 & 240 & 7.1\\
(f) & \ref{fig:RealProps}{\color{red}f} & Green & White & 88.9 & 240 & 7.1\\
(g) & \ref{fig:RealProps}{\color{red}g} & White & Black & 119.4 & 24 & 12.8\\
\bottomrule
\end{tabular}} \\[2pt]
\begin{tiny}
$^*$Case (c)'s light intensity shows High Dynamic Range scenario with illumination of the light part being 564lx and dark part being 2lx (See Fig. \ref{fig:RealProps}{\color{red}c}).
\end{tiny}
\end{table}

\subsubsection{Qualitative Evaluation on different real-world propellers}
In the final case study, we present qualitative results of \textit{EVPropNet} on different lighting scenarios, propeller sizes, propeller and background colors, $N_{\text{blades}}$, $r$ and angles. Fig. \ref{fig:RealProps} shows the qualitative results where the description of the scene is given in Table \ref{tab:PropScenes}. Notice how \textit{EVPropNet} can handle different real-world variations along with high dynamic range (Fig. \ref{fig:RealProps}{\color{red}c}, even a high-end DSLR cannot capture both shadows and highlights with it's 32dB of dynamic range but the event camera with it's 80dB can handle such a scene with ease), low contrast (Fig. \ref{fig:RealProps}{\color{red}d}) and low light with average intensity of 24lx (Fig. \ref{fig:RealProps}{\color{red}g}). 
\subsubsection{Quantitative Evaluation of applications}
We now present the results for both our applications and we call them experiment 1 for tracking and following and experiment 2 for landing. We define success as being able detect the quadrotor and to not completely loosing track for experiment 1, and for the quadrotor to be able to detect the other quadrotor and land on it successfully without collision for experiment 2. We average our results over 50 trails for each experiment and obtain a success rate of 92\% for experiment 1 and 90\% for experiment 2 (Vicon estimates for the trial shown in Fig. \ref{fig:Banner} are shown in corresponding sub-figures of Fig. \ref{fig:Vicon}). Commonly, the failure cases in experiment 1 happen when the larger quadrotor has a huge jerk that it moves outside the field of view of the camera.  The failure cases in experiment 2 happen due to the aerodynamic interference between the two quadrotors which makes the bottom quadrotor drift at the last moment.





\subsection{Analysis}
\label{subsec:Analysis}
\begin{figure}[t!]
    \centering
    \includegraphics[width=\columnwidth]{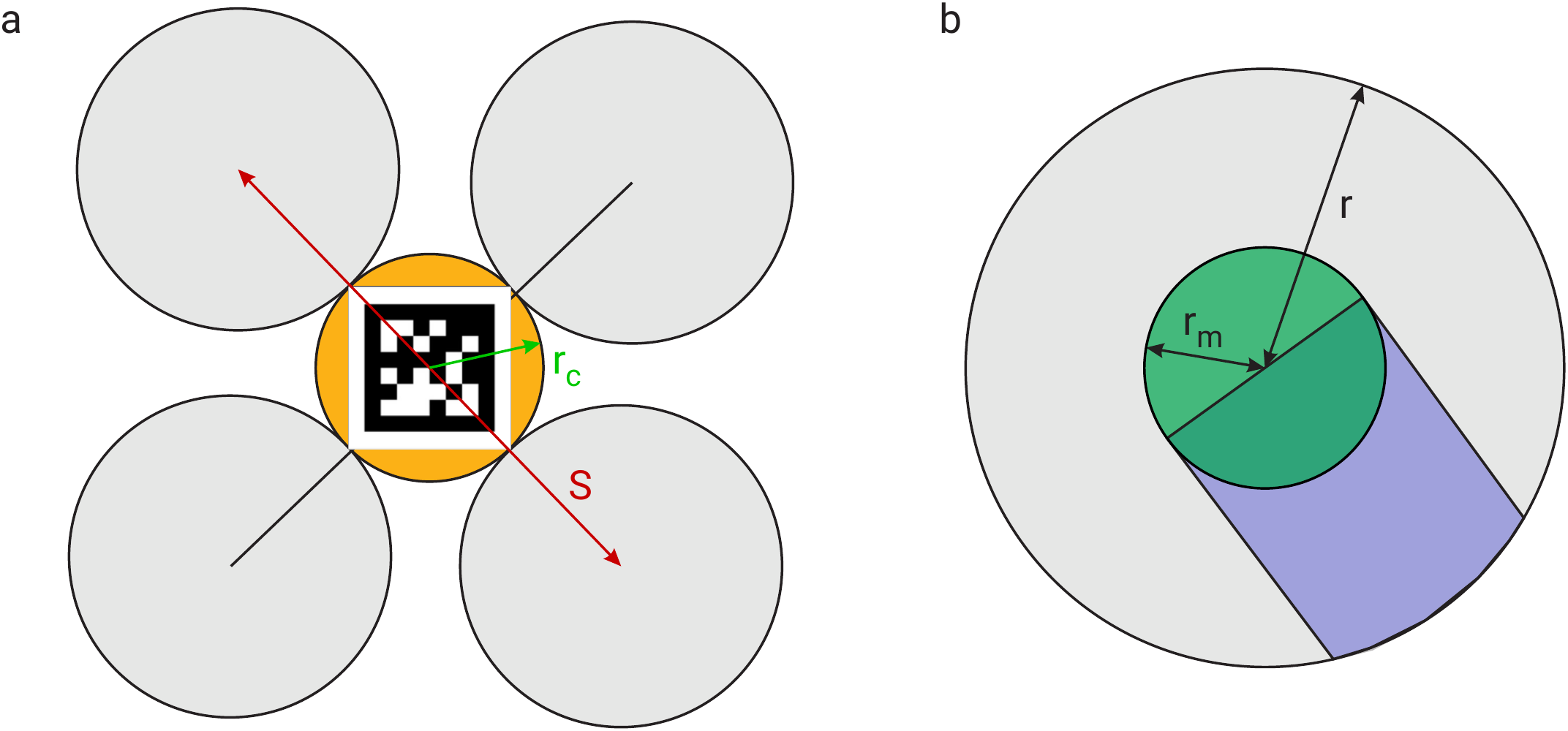}
    \caption{ (a) Simplified model of a quadrotor used to calculate area ratios of the propellers to that of the biggest square fiducial marker that can be fit in the center without obstruction, (b) Simplified arm and motor projection to compute amount of propeller occluded from generating events -- gray areas show where the propeller is visible and generates events, green area is occluded by the motor and blue area is occluded by the arm.}
    \label{fig:PropArea}
\end{figure}

We present analysis of three questions: 1. What is the ratio of visible area of the propellers to that of the largest square fiducial marker in the center? (Fig. \ref{fig:PropArea}) 2. How does DR vary with focal length $f$, real-world propeller radius $r$ and camera angle $\phi$ (angle around $X$ axis)? (Fig. \ref{fig:CoordinateFrames}{\color{red}d}). 3. What makes \textit{EVPropNet} generalize to the real-world without any fine-tuning or re-training?

To analyse the answer to the first question, we present a simplified geometric model of a multirotor (quadrotor shown in Fig. \ref{fig:PropArea}{\color{red}a}) where we are given a constraint on the drone's size $S$  (diagonal motor to motor length) and number of propellers on the drone $N_{\text{prop}}$. Let us say that the largest  fiducial marker that can fit in the center of the drone is inscribed in the circle of radius $r_c$. Also, we assume that the propeller does not generate events in the area in which is occluded by the arm and the motor. The motor radius is given by $r_m$ and the arm width is given by $2r_m$ (Fig. \ref{fig:PropArea}{\color{red}b}). The area of one non-occluded propeller $\mathcal{A}_{\text{prop}}$ (gray highlighted area in Fig. \ref{fig:PropArea}{\color{red}b}) is given by 

\begin{equation}
\label{eq:AreaEq1}
    \mathcal{A}_{\text{prop}} = r^2\left(\pi -\frac{\gamma}{2}\right) - \frac{\pi r_m^2}{2} - r_mr\cos\left( \frac{\gamma}{2}\right); \gamma = 2\sin^{-1}\left( \frac{r_m}{r}\right)
\end{equation}

Hence, the ratio for the area of the largest visible fiducial marker to that of a $N_{\text{prop}}$ propeller drone will be

\begin{equation}
\mathcal{A}_{\text{ratio}} = \frac{4N_{\text{prop}}\left( r^2\left( 2\pi - \gamma \right) - \pi r_m^2 - 2r_mr\cos\left( \frac{\gamma}{2}\right)\right)}{\left(S-2r\right)^2} 
\end{equation}

\begin{table}[t!]
\centering
\caption{\normalfont{$\mathcal{A}_{\text{ratio}}$} \textsc{for some common commercial drones.}}
\resizebox{0.9\columnwidth}{!}{
\label{tab:PropRatio}
\begin{tabular}{cccccc}
\toprule
Name & $S$ (mm) & $N_{\text{prop}}$ & $r$ (mm) & $r_m$ (mm) & $\mathcal{A}_{\text{ratio}}$\\
\midrule
 DJI Phantom 4 & 350 & 4 & 119.4 & 12 & 109.8\\
 QAV 210 X & 210 & 4 & 63.5 & 14.0 & 51.2 \\
 DJI Inspire 2 & 603 & 4 & 190 & 18.5 & 69.2\\ 
\bottomrule
\end{tabular}}
\end{table}

The value of $\mathcal{A}_{\text{ratio}}$ for some common commercially available drones are given in Table \ref{tab:PropRatio} (Recall, $N_{\text{prop}}$ is the number of propellers on the drone, $r$ is the propeller radius, $r_m$ is the motor radius, $\gamma$ is defined in Eq. \ref{eq:AreaEq1} and $S$ is the drone's diagonal motor to motor length). We clearly see that the probability of observing at-least one propeller (directly related to $\mathcal{A}_{\text{ratio}}$) is much higher than that of observing a fiducial marker in the middle, thereby reinforcing the motivation of our approach.

\begin{figure}[t!]
    \centering
    \includegraphics[width=\columnwidth]{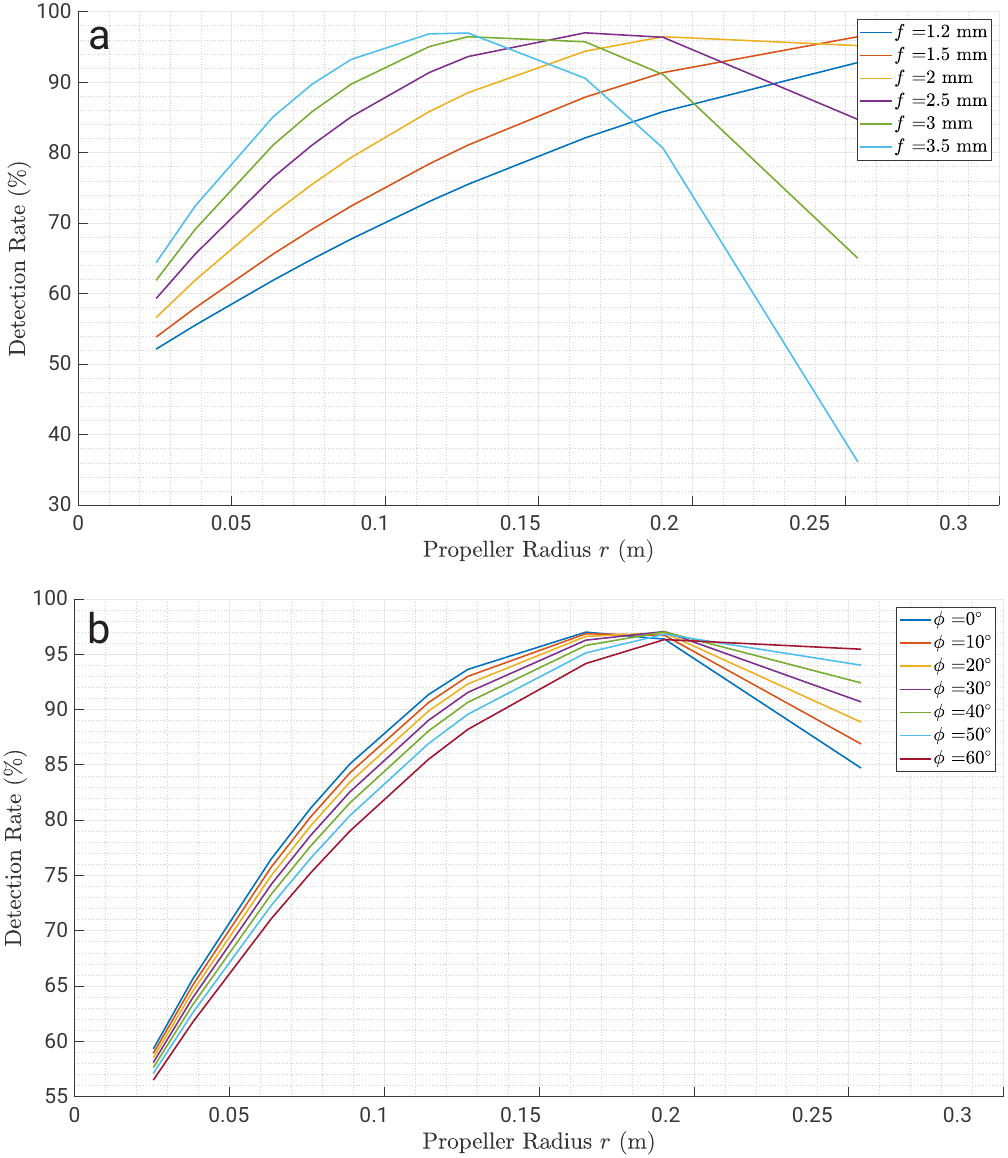}
    \caption{Variation of Detection Rate with variation in real-world propeller radius $r$ for different (a) Focal lengths $f$ with $\phi=0^\circ$, and (b) Camera Roll $\phi$ with $f=2.5$mm.}
    \label{fig:DRVsrvsphi}
\end{figure}

For the analysis of the second question, refer to Fig. \ref{fig:DRVsrvsphi}. We see that the DR of the propeller increases with an increase in real world propeller radius $r$ until it reaches a maximum and then decreases (Fig. \ref{fig:DRVsrvsphi}{\color{red}a}). This trend is observed since smaller propellers (small $r$) generate a small number of events leading to a low DR and increases with increase in number of events (directly correlated with $r$). However, with larger propellers the DR decreases as the number of false detections increase near the tip of the propeller. With a larger focal length (larger $f$), the curvature of the curve is larger since the relative projection area change (on the image plane) is more drastic. We can also observe a similar trend in Fig. \ref{fig:DRVsrvsphi}{\color{red}b} with change in angle $\phi$  (angle around $X^C$ axis in Fig. \ref{fig:CoordinateFrames}{\color{red}d}). Notice that the change in focal length affects the accuracy more significantly than the change in angle. Note that pitch $\theta$ has the similar effect to that of the roll $\phi$.

Finally, we speculate why our \textit{EVPropNet} generalizes to the real-world without any fine-tuning or re-training for different propellers.
\begin{itemize}
    \item  The data’s visual quality from simulation is similar to those obtained from recently developed event cameras both in-terms of noise and data-rate. (This is not simple with data from classical cameras due to the lack of photo-realism.)
    \item The errors in simulation (as compared to the real-world)  are lower when the integration time for creation of event frames are smaller (around 20ms maximum) as demonstrated by \cite{sanket2020evdodgenet}. 
\end{itemize}

\subsection{Implementation Considerations}
\label{subsec:Impl}
To speed-up the computation of our network when deployed on an aerial robot, we quantize our network to \texttt{Int8} and compile our network using \texttt{EdgeTPU} optimizations for deployment on the Google Coral USB Accelerator. To enable smooth compilation and high accuracy retention, we make our inputs take only valid \texttt{Int8} values as given below 
\begin{align}
    \mathpzc{E}_{\texttt{EdgeTPU}} &= \clamp\left(\mathpzc{E} \times 255 + 127 \vert 0, 255 \right)\\
    \clamp\left(x \vert a, b\right) &:= \max \left( b, \min \left(x,a \right)\right)
\end{align}

The labels $\hat{p}$ are modified as $\hat{p}_{\texttt{EdgeTPU}} = \lfloor \hat{p} \times 255 - 0.5 \rfloor$ and take integer values in $[0,255]$. 

Finally, when using an event camera with a high resolution at a high temporal sampling rate, the bottleneck of the system is the transfer speed between the event sensor and the compute module which are dictated by the combined throughput of processor, cache, transfer speeds of the primary and secondary memory. Such a bottleneck can cause data loss and data lag in the buffer. We mitigate this issue by using the NVIDIA TX2 which has a throughput of $\sim$440MBs$^{-1}$. 

\section{Conclusions}
\label{sec:Conc}
We presented a method to detect unmarked drones (multi-rotors) by detecting a ubiquitous part of their design -- the propellers. To enable detection of the propellers, we utilize the following fact: propellers rotate at high-speed and hence are generally the fastest moving parts on an image. We model the geometry of the propeller and use it to simulate the data from an event camera whose qualities of high temporal resolution, low latency and high dynamic range make it perfectly suited for detecting propellers. We then train our \textit{EVPropNet} deep network on this simulated data which generalizes directly to the real-world without any fine-tuning or re-training. We present two applications of detecting propellers on an unmarked drone: (a) tracking  and following  an unmarked  drone  and  (b)  landing  on  a  near-hover drone. As a parting thought, an active zoom camera would increase the distance range from where the drones could be detected and would make our method a viable for deployment in the wild.

\section*{Acknowledgement}
The support of the National Science Foundation under grants BCS 1824198 and OISE 2020624, the support of the Office of Naval Research under grant award N00014-17-1-2622, the Northrop Grumman Corporation and the Brin Family foundation are gratefully acknowledged. We also would like to thank Samsung for providing us with the event-based vision sensor used in this research.

\bibliographystyle{unsrt}
\bibliography{Ref}

\begin{thebibliography}{10}

\bibitem{Exploration}
Teodor Tomic et~al.
\newblock Toward a fully autonomous uav: Research platform for indoor and
  outdoor urban search and rescue.
\newblock {\em IEEE robotics \& automation magazine}, 19(3):46--56, 2012.

\bibitem{mcguire2017towards}
Kimberly McGuire, Mario Coppola, Christophe De~Wagter, and Guido de~Croon.
\newblock Towards autonomous navigation of multiple pocket-drones in real-world
  environments.
\newblock In {\em 2017 IEEE/RSJ International Conference on Intelligent Robots
  and Systems (IROS)}, pages 244--249. IEEE, 2017.

\bibitem{mcguire2019minimal}
KN~McGuire, Christophe De~Wagter, Karl Tuyls, HJ~Kappen, and Guido~CHE
  de~Croon.
\newblock Minimal navigation solution for a swarm of tiny flying robots to
  explore an unknown environment.
\newblock {\em Science Robotics}, 4(35), 2019.

\bibitem{sanket2020morpheyes}
Nitin~J Sanket, Chahat~Deep Singh, Varun Asthana, Cornelia Ferm{\"u}ller, and
  Yiannis Aloimonos.
\newblock Morpheyes: Variable baseline stereo for quadrotor navigation.
\newblock {\em arXiv preprint arXiv:2011.03077}, 2020.

\bibitem{Inspection}
Tolga {\"O}zaslan et~al.
\newblock {Inspection of penstocks and featureless tunnel-like environments
  using micro UAVs}.
\newblock In {\em Field and Service Robotics}, pages 123--136. Springer, 2015.

\bibitem{Mapping}
Friedrich Fraundorfer, Lionel Heng, Dominik Honegger, Gim~Hee Lee, Lorenz
  Meier, Petri Tanskanen, and Marc Pollefeys.
\newblock Vision-based autonomous mapping and exploration using a quadrotor
  mav.
\newblock In {\em 2012 IEEE/RSJ International Conference on Intelligent Robots
  and Systems}, pages 4557--4564. IEEE, 2012.

\bibitem{SearchAndRescue}
Nathan Michael et~al.
\newblock Collaborative mapping of an earthquake-damaged building via ground
  and aerial robots.
\newblock {\em Journal of Field Robotics}, 29(5):832--841, 2012.

\bibitem{GapFlyt}
Nitin~J Sanket, Chahat~Deep Singh, Kanishka Ganguly, Cornelia Ferm{\"u}ller,
  and Yiannis Aloimonos.
\newblock Gapflyt: Active vision based minimalist structure-less gap detection
  for quadrotor flight.
\newblock {\em IEEE Robotics and Automation Letters}, 3(4):2799--2806, 2018.

\bibitem{Mellinger2013}
Daniel Mellinger, Michael Shomin, Nathan Michael, and Vijay Kumar.
\newblock {\em Cooperative Grasping and Transport Using Multiple Quadrotors},
  pages 545--558.
\newblock Springer Berlin Heidelberg, Berlin, Heidelberg, 2013.

\bibitem{DroneMarket}
Zoran Valentak.
\newblock Drone market share analysis, 2018.

\bibitem{eventsurvey}
Guillermo Gallego, Tobi Delbr{\"{u}}ck, Garrick Orchard, Chiara Bartolozzi,
  Brian Taba, Andrea Censi, Stefan Leutenegger, Andrew~J. Davison, J{\"{o}}rg
  Conradt, Kostas Daniilidis, and Davide Scaramuzza.
\newblock Event-based vision: {A} survey.
\newblock {\em CoRR}, abs/1904.08405, 2019.

\bibitem{dvspaper}
P.~{Lichtsteiner}, C.~{Posch}, and T.~{Delbruck}.
\newblock A 128$\times$ 128 120 db 15 $\mu$s latency asynchronous temporal
  contrast vision sensor.
\newblock {\em IEEE Journal of Solid-State Circuits}, 43(2):566--576, 2008.

\bibitem{samsungdvs}
B.~{Son}, Y.~{Suh}, S.~{Kim}, H.~{Jung}, J.~{Kim}, C.~{Shin}, K.~{Park},
  K.~{Lee}, J.~{Park}, J.~{Woo}, Y.~{Roh}, H.~{Lee}, Y.~{Wang},
  I.~{Ovsiannikov}, and H.~{Ryu}.
\newblock 4.1 a 640×480 dynamic vision sensor with a 9µm pixel and 300meps
  address-event representation.
\newblock In {\em 2017 IEEE International Solid-State Circuits Conference
  (ISSCC)}, pages 66--67, 2017.

\bibitem{yolov3}
Joseph Redmon and Ali Farhadi.
\newblock Yolov3: An incremental improvement.
\newblock {\em arXiv}, 2018.

\bibitem{pawelczyk2020real}
Maciej~{\L} Pawe{\l}czyk and Marek Wojtyra.
\newblock Real world object detection dataset for quadcopter unmanned aerial
  vehicle detection.
\newblock {\em IEEE Access}, 8:174394--174409, 2020.

\bibitem{unlu2019deep}
Eren Unlu, Emmanuel Zenou, Nicolas Riviere, and Paul-Edouard Dupouy.
\newblock Deep learning-based strategies for the detection and tracking of
  drones using several cameras.
\newblock {\em IPSJ Transactions on Computer Vision and Applications},
  11(1):1--13, 2019.

\bibitem{schilling2020vision}
Fabian Schilling, Fabrizio Schiano, and Dario Floreano.
\newblock Vision-based drone flocking in outdoor environments.
\newblock {\em IEEE Robotics and Automation Letters}, 6(2):2954--2961, 2021.

\bibitem{walter2019uvdar}
Viktor Walter, Nicolas Staub, Antonio Franchi, and Martin Saska.
\newblock Uvdar system for visual relative localization with application to
  leader--follower formations of multirotor uavs.
\newblock {\em IEEE Robotics and Automation Letters}, 4(3):2637--2644, 2019.

\bibitem{mateos2020apriltags}
Luis~A. Mateos.
\newblock Apriltags 3d: Dynamic fiducial markers for robust pose estimation in
  highly reflective environments and indirect communication in swarm robotics,
  2020.

\bibitem{li2019modquad}
Guanrui Li, Bruno Gabrich, David Saldana, Jnaneshwar Das, Vijay Kumar, and Mark
  Yim.
\newblock Modquad-vi: A vision-based self-assembling modular quadrotor.
\newblock In {\em 2019 International Conference on Robotics and Automation
  (ICRA)}, pages 346--352. IEEE, 2019.

\bibitem{krogius2019flexible}
Maximilian Krogius, Acshi Haggenmiller, and Edwin Olson.
\newblock Flexible layouts for fiducial tags.
\newblock In {\em IROS}, pages 1898--1903, 2019.

\bibitem{calvet2016Detection}
Lilian Calvet, Pierre Gurdjos, Carsten Griwodz, and Simone Gasparini.
\newblock {Detection and Accurate Localization of Circular Fiducials under
  Highly Challenging Conditions}.
\newblock In {\em {Proceedings of the 2016 IEEE Conference on Computer Vision
  and Pattern Recognition (CVPR)}}, pages 562 -- 570, Las Vegas, United States,
  June 2016.

\bibitem{mitrokhin2018event}
Anton Mitrokhin, Cornelia Ferm{\"u}ller, Chethan Parameshwara, and Yiannis
  Aloimonos.
\newblock Event-based moving object detection and tracking.
\newblock In {\em 2018 IEEE/RSJ International Conference on Intelligent Robots
  and Systems (IROS)}, pages 1--9. IEEE, 2018.

\bibitem{stoffregen2019event}
Timo Stoffregen, Guillermo Gallego, Tom Drummond, Lindsay Kleeman, and Davide
  Scaramuzza.
\newblock Event-based motion segmentation by motion compensation.
\newblock In {\em Proceedings of the IEEE/CVF International Conference on
  Computer Vision}, pages 7244--7253, 2019.

\bibitem{parameshwara0}
Chethan~M Parameshwara, Nitin~J Sanket, Chahat~Deep Singh, Cornelia
  Ferm{\"u}ller, and Yiannis Aloimonos.
\newblock 0-mms: Zero-shot multi-motion segmentation with a monocular event
  camera.

\bibitem{sanket2020evdodgenet}
Nitin~J Sanket, Chethan~M Parameshwara, Chahat~Deep Singh, Ashwin~V
  Kuruttukulam, Cornelia Ferm{\"u}ller, Davide Scaramuzza, and Yiannis
  Aloimonos.
\newblock Evdodgenet: Deep dynamic obstacle dodging with event cameras.
\newblock In {\em 2020 IEEE International Conference on Robotics and Automation
  (ICRA)}, pages 10651--10657. IEEE, 2020.

\bibitem{falanga2020dynamic}
Davide Falanga, Kevin Kleber, and Davide Scaramuzza.
\newblock Dynamic obstacle avoidance for quadrotors with event cameras.
\newblock {\em Science Robotics}, 5(40), 2020.

\bibitem{carlton2018marine}
John Carlton.
\newblock {\em Marine propellers and propulsion}.
\newblock Butterworth-Heinemann, 2018.

\bibitem{ding2019efficient}
Wenchao Ding, Wenliang Gao, Kaixuan Wang, and Shaojie Shen.
\newblock An efficient b-spline-based kinodynamic replanning framework for
  quadrotors.
\newblock {\em IEEE Transactions on Robotics}, 35(6):1287--1306, 2019.

\bibitem{weisstein}
Eric~W. Weisstein.
\newblock B-spline. {From MathWorld---A Wolfram Web Resource}.

\bibitem{qin2000general}
Kaihuai Qin.
\newblock General matrix representations for b-splines.
\newblock {\em The Visual Computer}, 16(3-4):177--186, 2000.

\bibitem{MSCOCO}
Tsung-Yi Lin, Michael Maire, Serge Belongie, James Hays, Pietro Perona, Deva
  Ramanan, Piotr Doll{\'a}r, and C.~Lawrence Zitnick.
\newblock Microsoft coco: Common objects in context.
\newblock In David Fleet, Tomas Pajdla, Bernt Schiele, and Tinne Tuytelaars,
  editors, {\em Computer Vision -- ECCV 2014}, pages 740--755, Cham, 2014.
  Springer International Publishing.

\bibitem{ESIM}
Henri Rebecq, Daniel Gehrig, and Davide Scaramuzza.
\newblock {ESIM}: an open event camera simulator.
\newblock {\em Conf. on Robotics Learning (CoRL)}, October 2018.

\bibitem{he2016deep}
Kaiming He, Xiangyu Zhang, Shaoqing Ren, and Jian Sun.
\newblock Deep residual learning for image recognition.
\newblock In {\em Proceedings of the IEEE conference on computer vision and
  pattern recognition}, pages 770--778, 2016.

\bibitem{bianco2018benchmark}
Simone Bianco, Remi Cadene, Luigi Celona, and Paolo Napoletano.
\newblock Benchmark analysis of representative deep neural network
  architectures.
\newblock {\em IEEE Access}, 6:64270--64277, 2018.

\bibitem{sanket2020prgflow}
Nitin~J Sanket, Chahat~Deep Singh, Cornelia Ferm{\"u}ller, and Yiannis
  Aloimonos.
\newblock Prgflow: Benchmarking swap-aware unified deep visual inertial
  odometry.
\newblock {\em arXiv preprint arXiv:2006.06753}, 2020.

\bibitem{foehn2020alphapilot}
Philipp Foehn, Dario Brescianini, Elia Kaufmann, Titus Cieslewski, Mathias
  Gehrig, Manasi Muglikar, and Davide Scaramuzza.
\newblock {AlphaPilot: Autonomous Drone Racing}.
\newblock In {\em Proceedings of Robotics: Science and Systems}, Corvalis,
  Oregon, USA, July 2020.

\bibitem{jain2020flying}
Karan~P Jain and Mark~W Mueller.
\newblock Flying batteries: In-flight battery switching to increase multirotor
  flight time.
\newblock In {\em 2020 IEEE International Conference on Robotics and Automation
  (ICRA)}, pages 3510--3516. IEEE, 2020.

\bibitem{kalman1960new}
R.E. Kalman.
\newblock A new approach to linear filtering and prediction problems.
\newblock {\em Journal of Basic Engineering}, 82(1):35--45, 1960.

\bibitem{pfrommer2017penncosyvio}
Bernd Pfrommer, Nitin Sanket, Kostas Daniilidis, and Jonas Cleveland.
\newblock Penncosyvio: A challenging visual inertial odometry benchmark.
\newblock In {\em 2017 IEEE International Conference on Robotics and Automation
  (ICRA)}, pages 3847--3854. IEEE, 2017.

\end{thebibliography}

\end{document}